\documentclass[pmlr]{jmlr}% new name PMLR (Proceedings of Machine Learning)

\RequirePackage{graphicx}
\usepackage{booktabs}
\usepackage{multirow}
\usepackage{graphicx}
\usepackage{algorithmic}
\usepackage{longtable}% for long tables
\makeatletter
\def\set@curr@file#1{\def\@curr@file{#1}} %temp workaround for 2019 latex release
\makeatother
\usepackage[load-configurations=version-1]{siunitx} % newer version

\theorembodyfont{\upshape}
\theoremheaderfont{\scshape}
\theorempostheader{:}
\theoremsep{\newline}

\title[Knowledge-guided Casual Structure Search]{Optimizing Data-driven Causal Discovery Using Knowledge-guided Search}

\author{\Name{Uzma Hasan}
       \Email{uzmahasan@umbc.edu}\\ 
       \addr Causal AI Lab, Department of Information Systems\\
       University of Maryland, Baltimore County\\
       Baltimore, Maryland, USA 
       \AND
       \Name{Md Osman Gani}
       \Email{mogani@umbc.edu}\\ 
       \addr Causal AI Lab, Department of Information Systems\\
       University of Maryland, Baltimore County\\
       Baltimore, Maryland, USA} 

\begin{document}

\maketitle

\begin{abstract}
Learning causal relationships solely from observational data provides insufficient information about the underlying causal mechanism and the large search space of possible causal graphs. As a result, often the search space can grow exponentially, particularly for greedy algorithms that use a score-based approach to search the space of candidate causal graphs. Prior causal information such as the presence or absence of a causal edge can be leveraged to guide the score-based discovery process towards a more restricted and accurate search space. The healthcare domain is characterized by an abundance of prior knowledge from different sources such as numerous medical journals, databases like electronic health records (EHRs) and outcomes of clinical interventions. In this study, we present 
%a knowledge-guided score-based causal discovery (KGS) approach
a knowledge-guided causal structure search (KGS) approach
that uses observational data and structural priors such as causal edges as constraints to learn the causal graph. KGS is a novel application of knowledge constraints that can leverage any of the following prior edge information between any two variables: the presence of a directed edge, the absence of an edge, and the presence of an undirected edge. We extensively evaluate KGS across multiple settings in both synthetic and benchmark real-world datasets. We also test the performance of KGS on a real-life healthcare application related to oxygen therapy treatment. To obtain causal priors, we conduct an analysis that leverage GPT-4 to retrieve information about any causal edge available in relevant literature. Our experimental results demonstrate that structural priors of any type and amount are helpful and guide the search process toward improved performance and optimized causal discovery. This guided strategy ensures that the discovered edges support the established causal knowledge, which improves the trustworthiness of the findings while simultaneously expediting the search process. It also enables a more targeted exploration of causal mechanisms, which can result in more effective and personalized healthcare solutions.
\end{abstract}

\section{Introduction}

\label{sec:intro}
Causal discovery (CD) deals with unfolding the causal relationships among entities in a system. The causes and their corresponding effects are inferred from data, and represented using a \textit{causal graph }that consists of nodes representing domain variables, and arrows indicating the direction of the relationship between those variables (\cite{hasan2023survey}). Causal graphs are often represented using a Directed Acyclic Graph (DAG) to visually depict the causal relationships between variables. These graphs help clinicians and researchers identify which factors influence patient outcomes, thereby facilitating more effective interventions and personalized treatment plans (\cite{yang2013causal}, \cite{sanchez2022causal}). For example, causal graphs can illustrate how lifestyle choices, genetic conditions, and environmental factors contribute to the development of a particular disease. This understanding is crucial for developing preventive measures and targeting therapies that address the root causes of diseases rather than just the symptoms. There exist different approaches to discover the causal structure from data among which the constraint-based (\cite{PC}, \cite{rfci}) and score-based approaches (\cite{chickering2002optimal}, \cite{SGES}) are the most prominent ones. Constraint-based approaches perform a series of conditional independence tests to find the causal relationships from data. Whereas, the score-based approaches search over the space of possible graphs to find the graph that best describes the data. Often, these approaches use a search process such as greedy, A*, and other heuristic searches (\cite{domainA_star}) combined with a score function such as BIC, AIC, MDL, and BDeu (\cite{meek2013causal}) to score all the candidate graphs. The final outcome is one or more causal graphs with the highest score. 

Among the score-based approaches, a commonly used approach is the Greedy Equivalence Search (GES) (\cite{chickering2002optimal}) that searches over the space of equivalence classes of causal graphs. 
% It is an iterative process that starts with an empty network and greedily adds edges one at a time until it reaches the local maximum. Then it continues with a series of greedy deletions of edges until the score keeps improving. 
Although it is widely used, there are some major disadvantages of this search technique. First, the search space becomes exponential when the number of possible candidate states increases rapidly with the increase in the number of variables (
%\cite{SGES}, 
\cite{SE-GES}). This is because it considers all possible combinations of candidate search states. Hence, a large number of combinations must be considered as the number of variables keeps growing
%, resulting in an exponential growth of the search space 
(\cite{SGES}). Even for sparse graphs, where the edge denstities are smaller, the search space is large enough to negatively impact the efficiency and performance of the algorithm. Second, the score needs to be computed for each possible graph which makes it computationally expensive (\cite{SGES}, \cite{SE-GES}). The situation worsens while traversing the search space of densely-connected graphs. Because it drastically increases the number of times the algorithm needs to compute the score as it considers almost every node in the respective graph subspace %(\cite{SE-GES})
. Furthermore, the search process is typically repeated multiple times, which adds to the overall cost. 

%\textcolor{red}{Add we have results from literature/RCTs/studies}
The healthcare domain is fortunately rich in prior knowledge. There is often prior knowledge about the causal connections between some of the variables in various cases.
% Often in many domains, there exists some prior information about the causal relationships between some of the variables. 
Such causal priors may be obtained from multiple sources such as randomized control trials (RCTs), disease registries, EHRs, and systematic reviews. Other sources encompass a wide array of established scientific research, clinical expertise, and real-world data. There is lack of efficient approaches that integrate causal priors in a comprehensive manner during causal discovery. Currently, most of the existing causal discovery approaches are solely data-driven, rely heavily on the data samples, and do not always consider the available causal knowledge (\cite{domainA_star}). However, researchers are now getting interested to augment structure learning with known information about causal edges and study their effectiveness to mitigate various practical challenges of causal discovery solely from observational data (\cite{hasan2022kcrl}, \cite{k-notears}). %GES may also be benefited when prior knowledge about some of the causal relationships is used in the form of additional constraints \cite{chickering2002optimal}.  
To address the challenges in greedy score-based approaches, existing causal information can be efficiently used during the discovery process. 
Knowledge constraints may restrict the algorithmic search space by shifting the focus to a smaller set of potential solutions, thereby, providing a better understanding of the context in which the search is being performed. This may lead towards early convergence resulting in a lower computational cost as well as reducing the score computations and number of search states that need to be explored. This guided approach not only makes the search process more efficient but also enhances the reliability of new findings by ensuring they support the established causal knowledge. It enables a more targeted exploration of causal mechanisms, which can lead to more effective and personalized healthcare solutions.

To the best of our knowledge, there is no comprehensive study focussing on the application and impact of leveraging the existing causal knowledge in a greedy score-based causal discovery approach. Therefore, in this work, we present an approach \textbf{K}nowledge-\textbf{G}uided Causal \textbf{S}tructure Search (KGS), which leverages knowledge constraints in a systematic way and study how these additional constraints may help to guide the search process. We consider three types of causal-edge constraints: (i) Directed edges ($\rightarrow$), (ii) Forbidden edges ($\not\to$), and (iii) Undecided edges (--). These types of prior information about the causal relationships are often available in healthcare \cite{}. Here, \textit{directed} edge and \textit{forbidden} edge refers to the \textit{existence} or \textit{absence} of a causal relationship respectively, and \textit{undecided} edge refers to the presence of a causal relationship whose \textit{direction is unknown}. 

We evaluate the performance of KGS with the three types of edges as well as a combination of all the edges on three synthetic and three benchmark real datasets. We also compare KGS with baseline algorithms including PC (\cite{PC}, GES (\cite{chickering2002optimal}) and Direct-LiNGAM (\cite{directlingam}) and also, present a comparative analysis of how different types of structural priors influence the search process in terms of both graph discovery and computational accuracy. We further study how varying the amount of leveraged knowledge affects the performance of the graph discovery. Lastly, we perform another analysis for the efficient extraction of causal priors from academic literature by mining relevant research papers using the GPT-4 model. We observe that GPT-4 can be used carefully to efficiently extract causal statements from relevant literature. This information can be leveraged 
%to automate the preliminary 
at different
stages of the causal discovery process. To summarize, we aim to answer the following questions in this study: 1) How do knowledge constraints affect the learned graph’s accuracy?, 2) Which type of knowledge constraint is the most effective?, 3) How does varying the amount of knowledge influence the performance?, and 4) Can we extract some causal priors from (healthcare) literature by leveraging LLMs and utilize those for causal discovery? Our \textbf{main contributions} are summarized below:
%4) Do knowledge constraints helps to achieve an early convergence to the optimal causal graph and improves computational efficiency?  
\begin{itemize}
    \item We present KGS, a novel application of causal constraints that leverages available information about different types of edges (directed, forbidden, and undecided) during causal discovery.
    \item  We demonstrate how the search space as well as scoring candidate graphs can be reduced when different edge constraints are leveraged during a search over equivalence classes of causal graphs.
    \item Through an extensive set of experiments in both synthetic and benchmark real-world healthcare settings, we show how different types of prior knowledge can impact the discovery process of graphs of varied densities (small, medium, and large networks). We also study the influence of different proportion of knowledge constraints on the structure recovery through a set of experiments.
    \item We also validate KGS on an actual clinical problem pertaining to oxygen therapy in the intensive care unit. The clinical application is related to a vital and significant healthcare problem that can be utilized for patients suffering from severe acute respiratory syndrome coronavirus-2.
    \item We further do an analysis by leveraging large language models (LLMs) to extract causal priors from relevant literature evidence. Particularly, we use the GPT-4 model to extract the causal relation between pairs of variables from relevant healthcare research articles.
\end{itemize}

\subsection*{Generalizable Insights about Machine Learning in the Context of Healthcare}
Causal structures play a crucial role in machine learning by enhancing model understanding, robustness, and predictability, especially in real-world applications where understanding the underlying mechanisms of data generation is important. Causal structure learning solely from observational data is often challenging due to the lack of large amounts of data to reach statistical significance. Therefore, relying solely on observational data without any interventions or prior knowledge can severely limit the reliability of the derived causal graphs, making them less useful for decision-making purposes. Fortunately, we often possess some prior knowledge or available experimental evidence. Prior knowledge sources in healthcare include a wide range of established scientific research, systematic reviews, clinical trials, epidemiological studies, and expert viewpoints. 
%documented in medical literature, such as journals, textbooks, and systematic reviews. 
With the growing usage of LLMs, causal relationships retrieved by such models can be another valauble source of prior knowledge that can significantly aid in the process of causal discovery. The true potential of structure learning algorithms and data can be better realized only when some form of prior knowledge is applied. In this study, we discuss an efficient way to incorporate different types of prior knowledge during causal discovery. Our goal is to optimize the search process and reduce the exponential search space of candidate graphs by leveraging different edge constraints during searching over candidate causal graphs. The approach can be utilized in a broad range of healthcare problems whenever some prior knowledge is available. The obtained causal graphs can lead to more robust, accurate, and valid conclusions, which are crucial for effective decision-making and policy formulation in healthcare.

\section{Related Work}

Conventional causal discovery approaches include \emph{Constraint-based} approaches that tests for conditional independencies among variables (\cite{PC}, \cite{anytime}) and \emph{Score-based} approaches that scores candidate causal graphs to find the one that best fits the data (\cite{chickering2002optimal}, \cite{SGES}). \textit{Hybrid }methods leverage conditional independence tests to learn the skeleton graph and prune a large portion of the search space combined with a score-based search to find the causal structure (\cite{mmhc}, \cite{hcm}). Other common approaches include \emph{function-based} methods that represent variables as a function of its parents and an independent noise term (\cite{LiNGAM},\cite{anm}). Some recent popular \emph{gradient-based} approaches use neural networks and a modified definition of the acyclicity constraint (\cite{notears}, \cite{dag-gnn}, \cite{gran-dag} etc.) that transforms the combinatorial search to a continuous optimization search. The conventional approaches do not consider any knowledge constraints in the search process. However, multiple studies (\cite{fenton2018risk}, \cite{k-notears}) mention the importance of considering existing causal information when learning causal representation models. \cite{meek2013causal} was one of the earliest to suggest orientation rules for incorporating prior knowledge in constraint-based causal discovery approaches. 

In recent years, there has been an increasing amount of interest in the incorporation of knowledge into causal structure learning (\cite{CPDAGwithBG}). \cite{amirkhani2016exploiting} investigated the influence of the same kinds of prior probability on edge inclusion or exclusion from numerous experts which were applied to two variants of the Hill Climbing algorithm. \cite{tiredFCI} incorporates tiered background knowledge in the constraint-based FCI algorithm where they mention that prior knowledge allows for the identification of additional causal links. \cite{ida} explores the task of estimating all potential causal effects from observational data using direct and non ancestral causal information. Recently, \cite {hasan2022kcrl} proposed a generalized framework that uses prior knowledge as constraints to penalize an RL-based search algorithm which outputs the best rewarded causal graph. \cite{domainA_star} shows that even small amounts of prior knowledge can speed up and improve the performance of A*-based causal discovery. \cite{k-notears} studies the impact of expert causal knowledge in a continuous optimization-based causal algorithm and suggests that the practitioners should consider utilizing prior knowledge whenever available. \cite{gani2023structural} presents a causal discovery framework that considers advice from experts once the causal graphs are produced by the majority voting of some existing algorithms. Although some approaches have explored the incorporation of prior knowledge in multiple ways, to the best of our knowledge there is no approach that studies the impact and application of different types of edge constraints in a greedy score-based causal discovery search. %Therefore, it would be interesting to see the implications of different types of knowledge constraints (edge information) in a greedy score-based approach.

%\textbf{ADD SCM-OT paper}

\section{Methodology}

In this section, we 
%introduce our approach \textbf{K}nowledge \textbf{G}uided Causal \textbf{S}tructure Search abbreviated as KGS, 
discuss KGS
that uses a set of user-defined knowledge constraints to search for a causal graph that best fits the data and knowledge constraints. %The constraints allow KGS to complete the search process using a reduced set of modification operators. 
KGS primarily works in three steps: \textit{(i) knowledge organization, (ii) addition of edges and (iii) deletion of edges } which are discussed below and summarized in Algorithm~\ref{algo1}.

\paragraph{Types of knowledge constraints:}
We consider the following types of constraints (causal edges) between the nodes of a causal graph $G$:
%i) Presence of an edge with direction known, ii) Absence of an edge and iii) Presence of an edge with direction unknown. They are discussed in detail below.\\

(i) \textit{\textbf{Directed edge (d-edge)}}: The existence of a directed edge (→) from node $X_{i}$ (cause) to node $X_{j}$ (effect). This signifies that these nodes are causally related and $X_{i}$ is the cause of the effect $X_{j}$.

(ii) \textit{\textbf{Forbidden edge (f-edge)}}: The absence of an edge or causal link ($\not\to$) between two nodes $X_{i}$ and $X_{j}$. It signifies the non-existence of a causal relationship between the nodes.

(iii) \textit{\textbf{Undecided edge (u-edge)}}: These are the type of edges whose existence is known, however, their directions are unknown. It signifies the presence of an undirected edge (--) between two nodes $X_{i}$ and $X_{j}$ without any information about the direction of causality.

\paragraph{Assumptions:} In this study, we make the following assumptions. \textit{First}, we assume that there are no hidden variables or selection bias. 
%We also make some assumptions about the knowledge constraints. 
\textit{Second}, we assume that the knowledge constraints are 100\% true without any bias or error. \textit{Third}, there can't be any conflict among the knowledge constraints. That is the same constraint can't fall into multiple categories. An edge can't be directed and undecided at the same time.

\paragraph{Knowledge set construction:} We briefly discuss the different ways of obtaining causal priors to construct a knowledge set that can be used during causal discovery.
\begin{itemize}
    \item Literature evidence: Causal priors can be efficiently extracted from literature evidence through a systematic review of academic journals, conference papers, articles, textbooks, and case studies.
    \item Expert viewpoint: By interacting with domain experts, researchers can gain insightful knowledge about causal interactions that are based on observational skill and practical experience.
    \item LLM-based retrieval: Prior knowledge can be extracted from Large Language Models (LLMs) by querying them with specific prompts about causal relationships within a domain. We perform an analysis where we leverage the GPT-4 model to extract the causal relation between pairs of variables from relevant healthcare research papers (details in subsection~\ref{LLM-based-ext}).
    
\end{itemize}

\noindent
\paragraph{Steps in KGS: }We discuss below in detail the steps involved in KGS. 
\vspace{5pt}

\textbf{Step-1}: \textit{\textbf{Knowledge Organization}}. In this step, a knowledge set \textit{K }is formulated using the available prior causal edges $e_{k}$. The knowledge set is basically a $d \times d$ matrix with rows-column denoting the variables and entries denoting their causal relation (prior edges). As per our assumption of bias-free knowledge, only edges that are $100\%$ reliable must be used in this process. Depending on the type of knowledge (directed, forbidden or undecided edge), a $d \times d$ matrix is formed where the entries with ‘1’ indicate a directed edge, ‘2’ for forbidden edge and ‘3’ for undecided edge. Other entries denoted by ‘0’ indicate that there is no information about that edge or relation. KGS can be used with any one of the three types of knowledge or even with a combination of all of them. 

\textbf{Step-2}: \textit{\textbf{Edge Addition Phase}}. To improve the search where edges that provide the current best gain are greedily added, instead of starting with a no-edge model, we start with an initial graph that consists of the edges $e_{k}$ present in the knowledge set \textit{K}. KGS tries to restrict the initial vast set of insert operators (candidate edges than can be added) based on the \textit{directed} or \textit{undecided} edges available in \textit{K.} This helps in reducing the initial large number of candidate states that were needed to be examined when there is no knowledge constraint. Furthermore, the insert operators that conflict with the \textit{forbidden} edges in \textit{K} are completely ignored during this phase and only $K$-consistent insert operators $I$ are used. This ensures that for any two nodes ($X_{i}$, $X_{j}$) with a $\not\to$ edge in $K$, no unnecessary search is conducted to add an edge between them. This causes the reduction of many search states that might have been explored previously if no knowledge constraints were leveraged. Precisely, it rules out all of the candidate graphs which violate these edges. To score the candidate graphs we use the Bayesain Information Criterion (BIC). Going forward in each iteration we choose the candidate graph that gives the highest increment in score, i.e. the graph whose score $S^{'}$ is better than the current best score $S$, and update the model with the changes done. When there is no further refinement of the score ($S^{'}$ $<$ $S$), this phase stops and the search continues further to the next step.

\begin{algorithm}
    \caption{Knowledge Guided Causal Structure Search (KGS)}
    \small
    \SetAlgoLined
    \DontPrintSemicolon
    \textbf{Input:} Data $D$, Prior causal edges $E_{k}$.
    
    \textbf{Output:} Causal graph $G$.
    
    Initialize vertex set $V = \{1, ..., d\}$ and edge set $E = \emptyset$ % define an empty edge set
    
    Construct knowledge set $K \gets E_{k}$ $\quad$ $\quad$ %//Initialize matrix $K$ with edges $e_{k}$
    
    $E = E \cup K$
    
    %Construct a CPDAG $C$ that has $K$-consistent edges
    $G$ ($V$, $E$) $\leftarrow$ a graph with $v \in V$ and $e \in E$
    
    $S$ $\leftarrow$ $S_{BIC} (G)$  $\quad$ //Initialize score
    
    $\quad$ //edge addition starts
    
    \For  {each $i$ $\in$ $V$}{ 
        \For  {each $j$ $\in$ $V$}{ 
            $I$ $\leftarrow$ $K$-consistent insert operators
            
            \If{$i \not\to j \notin K$}{
            
            $G^{'} \leftarrow G \cup (i \rightarrow j)$\\
            % Compute $S_{BIC} (G)$ for $i \rightarrow j$ $\in$ $G$
            Compute $S^{'} \leftarrow S_{BIC} (G^{'})$\\ 
               \If{$S^{'} > S$}{
               % Add $i \rightarrow j$ to $C$               
               $G \leftarrow G^{'}$\\
               $S$ $\leftarrow$ $S^{'}$
               }
            
            }

               % \Else{Do not add $i \rightarrow j$ to $G$}
            
            % \If {$f$ edges $\in$ $K$}{
            %     % $PS_{i} \gets $($PS^{temp}_{i} - PS^{loss}_{i}$)
            %     %Generate $K$-consistent insert operators $I$ 

            %     Add $i \rightarrow j$ to $G$ if $S_{BIC} (G) > S$ and $\notin$ $I$
                
            %     %Add edge if it improves $G$ and is consistent with $I$ 
            % }
            % \Else {
            %   % Add edge if it improves $G$
               
            %   Add $i \rightarrow j$ to $G$ if $S_{BIC} (G) > S$
            % }
        }
    }    
    $\quad$ //edge deletion starts

    \For  {each $i$ $\in$ $V$} { 
        \For  {each $j$ $\in$ $V$}{ 
            $D$ $\leftarrow$ $K$-consistent delete operators
            
            \If{$i \rightarrow j$ or $i - j$ $\notin$ $K$}{ 
            
            $G^{'} \leftarrow G \cup (i \not\to j)$
            
            Compute  $S^{'} \leftarrow S_{BIC} (G^{'})$
               
               \If{$S^{'} > S$}{
               $G \leftarrow G^{'}$
               
               $S$ $\leftarrow$ $S^{'}$
               }
            }

               %\Else{Do not remove $i \rightarrow j$ from $C$}
            
            % \If {$f$ edges $\in$ $K$}{
            %     % $PS_{i} \gets $($PS^{temp}_{i} - PS^{loss}_{i}$)
            %     %Generate $K$-consistent insert operators $I$ 

            %     Add $i \rightarrow j$ to $C$ if $S_{BIC} (C) > S$ and $\notin$ $I$
                
            %     %Add edge if it improves $C$ and is consistent with $I$ 
            % }
            % \Else {
            %   % Add edge if it improves $C$
               
            %   Add $i \rightarrow j$ to $C$ if $S_{BIC} (C) > S$
            % }
        }
    }

    return $G$
    
    \label{algo1}
\end{algorithm}
%\paragraph{Working Procedure} 
\noindent

\textbf{Step-3}: \textit{\textbf{Edge Deletion Phase}}. To further refine the graph obtained in the previous phase, we iteratively delete edges based on the score and remove the delete operators (candidate edges that can be deleted) that conflict with the knowledge set to restrict the unnecessary search states. Particularly, the operators that contradict with the \textit{directed} or \textit{undecided} edges present in $K$ are ruled out and only $K$-consistent delete operators $D$ are used. 
%to make sure that any prior edges ($\rightarrow$ or $--$ ) between nodes ($X_{i}$, $X_{j}$) are not deleted. 
Similar to the earlier step, the removal of an edge is allowed only when it improves the score. This process terminates if there is no further improvement in score. Finally, the output is the causal graph obtained in the final iteration. The \textbf{time complexity} of the Algorithm~\ref{algo1} is $O(n^{2})$ in the worst case since both the loops are nested loops.

\begin{figure}[!htb]
  \centering
  \includegraphics[width=0.65\linewidth]{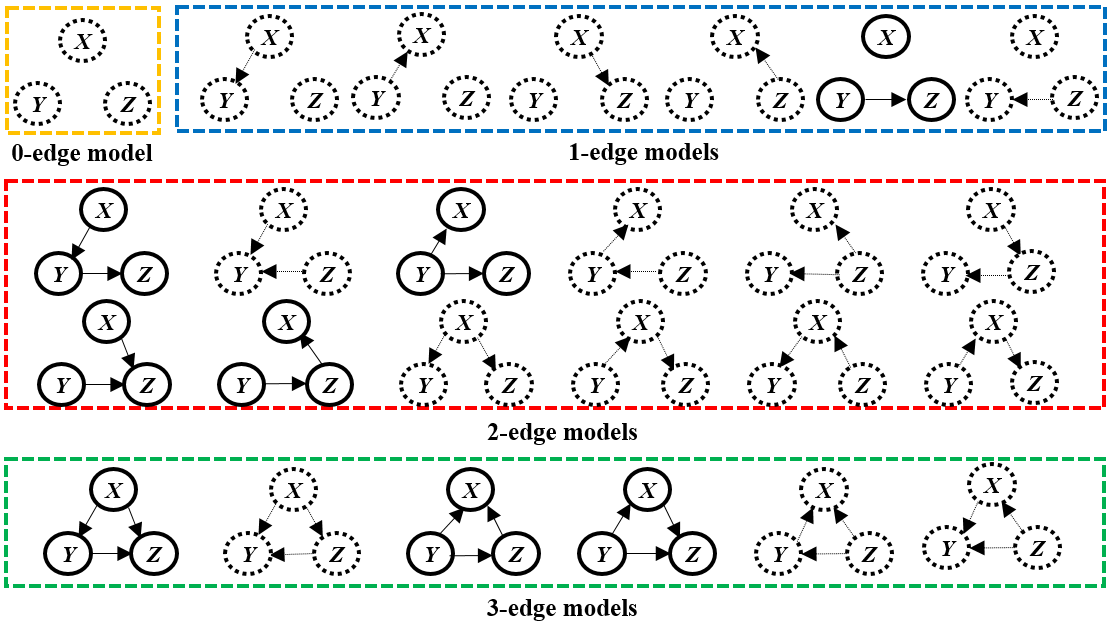} 
  \caption{Reduction of the initial search space of a 3-node graph due to leveraging a single causal knowledge (directed edge). Here, the dotted graphs can be ignored during the search process based on the prior knowledge $Y \rightarrow Z$ and the bolded graphs are the ones to be considered during the search.}\label{fig1}
\end{figure}

\paragraph{Computational benefits:}
A potential practical problem in greedy score-based search is the vast search space that grows exponentially with the increasing number of variables. %Unlike many other search algorithms, GES does not make use of any heuristics or domain-specific knowledge. The algorithm simply looks at every possible combination of rules and evaluates them based on the number of equivalences they produce. As the number of rules and variables increases, the search space grows exponentially. 
However, with the use of any available knowledge constraints, this vast growth in the search space can be controlled to a good degree. The second and third phases in KGS take into account the user-defined knowledge constraints which influences in shifting the overall search strategy to a more accurate trajectory. Let us consider the case of a 3 variable setting in Figure~\ref{fig1}. If we start the search with zero knowledge constraints (empty graph), then the space of candidate DAGs consists of a total 25 possible DAGs to choose from. Now, we suppose we are aware of the prior knowledge that a directed edge \textit{Y→Z} exists for the 3-node graph with variables $X$, $Y$ and $Z$. Thus, with this prior information, KGS starts with a single-edge model instead of the zero-edge model, and the space of all candidate DAGs are reduced to 8 possible states only. Since we are assuming constraints that are fully accurate, we can prune away the contradictory DAGs, leaving only DAGs that are consistent with the prior knowledge. Thus, it needs to score a much lower number of graphs than earlier due to the guidance of a single edge only. Also, it need not perform expensive score function evaluations for those DAGs that have been pruned away, which may result in significant computational gains. 
% Since in the worst case scenario, the number of search states that GES needs to evaluate can be exponential to the number of nodes (\cite{SGES}), it is computationally beneficial even if a little amount of knowledge is used to guide or restrict the vast space. 
Enormous savings can be gained by not generating candidates that we are aware of being invalid (\cite{chickering2002optimal}). Studies such as \cite{chickering2002optimal} suggests that we can prune neighbors heuristically to make the search practical. Leveraging existing knowledge about the causal edges can work as a heuristic because it may help to restrict the search only through candidate DAGs that are knowledge consistent.

\section{Experiments}\label{Experiments}

We conduct a set of comprehensive experimental evaluations to demonstrate the effectiveness of our proposed approach. We study the impact of directed, forbidden and undecided edges individually as well as the impact of a combination of the three types of edges. We report their performance by comparing the estimated graphs with their individual ground truth graphs  $G_{T}$. 
% To distinguish the experiments with different types of edges, we name the experiments as follows. 
We distinguish the experiments with directed, forbidden and undecided edges by naming them as  KGS-d, KGS-f and KGS-u respectively. Furthermore, the experiment that uses a combination of all of these edges is named KGS-c.  %Therefore, KGS-c is a combination of the steps involved in KGS-d, KGS-f and KGS-u. 
We compare the performance of KGS-d, KGS-f, KGS-u and KGS-c with three baseline approaches namely PC, GES and dLiNGAM on synthetic and benchmark real datasets as well as on a practical healthcare application. %We experiment with datasets having discrete as well as continuous variables. 
For each experiment, we randomly sample a fixed number of edge constraints from the ground-truth network $G_{T}$ (precisely 25\%). As a score function, we use the Bayesain Information Criterion (BIC) as it is a standard one as well as the most commonly used score function for causal discovery.

\paragraph{Metrics:} For performance evaluation, we use three common causal discovery metrics that represent the accuracy or closeness of the estimated causal graphs compared to the ground truth: \textit{(i) Structural Hamming Distance (SHD)} which denotes the total number of edge additions, deletions and reversals required to transform the estimated graph $G$ to the ground-truth DAG $G_{T}$, \textit{(ii) True Positive Rate (TPR)} that denotes the ratio of discovered true edges with respect to the total number of edges discovered and \textit{(iii) False Discovery Rate (FDR)} which denotes the proportion of the estimated false edges. \textit{Lower} the values of SHD and FDR, and a \textit{higher} TPR resembles a better causal graph. 
%We also report the run time (in seconds) of each experiment to see if there is any improvement in terms of computational efficiency. 

%The reported metric values are the mean values (averaged over 5 seeds/runs).

\paragraph{Setup:} The experiments are conducted on a 4-core Intel Core i5 1.60 GHz CPU cluster, with each process having access to 4 GB RAM. The implementations of the PC and dLiNGAM algorithms have been adopted from the gCastle (\cite{gcastle} repository. For GES, we used the publicly available python implementation in the \href{https://github.com/py-why/causal-learn}{causal-learn package}. \textit{The code implementation and experimental datasets of KGS have been uploaded as supplementary materials. Those will be made public after the blind review period is over.}
%We implemented KGS by extending the publicly available python implementation of GES in the \href{https://github.com/py-why/causal-learn}{causal-learn package}. 

\subsection{Synthetic Data}\label{sec:figures}
% Figures should go in the \texttt{figure} environment and be centered therein.
% The caption should go below the figure.
% Use \verb|\includegraphics| for external graphics files but omit the file extension.
% Supported formats are \textsf{pdf} (preferred for vector drawings and diagrams), \textsf{png} (preferred for screenshots), and \textsf{jpeg} (preferred for photographs).
% Do not use \verb|\epsfig| or \verb|\psfig|.
% If you want to scale the image, it is better to use a fraction of the line width rather than an explicit length.
% For example, see Figure~\ref{fig:pitt}.

% Do not use \verb|\graphicspath|.
% If the images are contained in a subdirectory, specify this when you include the image, for example \verb|\includegraphics{figures/mypic}|.

To evaluate the performance on synthetic datasets, we employ a similar experimental setup as \cite{notears} to generate random graphs with number of nodes $d = {10, 40, 100}$ using the  Erdős–Rényi (ER) model. The number of nodes are chosen as such to ensure that our approach is tested against networks of all sizes:\textit{ small, medium and large}. Each graph has an edge density of $e$ = 2$d$ where uniform random weights $W$ are assigned to the edges. Data $X$ is generated by taking $n = 1000$ i.i.d samples from the linear structural equation model (SEM) $X = W^{T}X + z$, where $z$ is a Gaussian noise term.

\begin{table*}[!h]
\small
\centering
\caption{Performance on the synthetic datasets with the best results for each dataset boldfaced. KGS-d has the best SHD and TPR for all the datasets. KGS-d also has the best FDR in case of two out of three datasets and PC achieved the best FDR for the dataset with small network size.}
\label{table1}
\begin{tabular}{ccccccccccccc}
 \toprule
{\color[HTML]{000000} } & \multicolumn{3}{c}{{\color[HTML]{000000} SHD $\downarrow$ }} & \multicolumn{3}{c}{{\color[HTML]{000000} TPR $\uparrow$ }} & \multicolumn{3}{c}{{\color[HTML]{000000} FDR $\downarrow$ }} \\ \cmidrule(lr){2-4}\cmidrule(lr){5-7}\cmidrule(lr){8-10}
{\color[HTML]{000000} } & {\color[HTML]{000000} \textit{d-10}} & {\color[HTML]{000000} \textit{d-40}} & {\color[HTML]{000000} \textit{d-100}} & {\color[HTML]{000000} \textit{d-10}} & {\color[HTML]{000000} \textit{d-40}} & {\color[HTML]{000000} \textit{d-100}} & {\color[HTML]{000000} \textit{d-10}} & {\color[HTML]{000000} \textit{d-40}} & {\color[HTML]{000000} \textit{d-100}}  \\ \midrule
PC & 13 & 74 & 187 & 0.42 & 0.5 & 0.47 & \textbf{0.47} & 0.49 & 0.5\\
{\color[HTML]{000000} GES} & {\color[HTML]{000000} 6} & {\color[HTML]{000000} 60} & {\color[HTML]{000000} 107} & {\color[HTML]{000000} 0.67} & {\color[HTML]{000000} 0.81} & {\color[HTML]{000000} 0.84} & {\color[HTML]{000000} 0.71} & {\color[HTML]{000000} 0.5} & {\color[HTML]{000000} 0.41} \\
dLiNGAM & 18 & 169 & 246 & 0.37 & 0.4 & 0.6 & 0.53 & 0.79 & 0.6\\
%NOTEARS & 3 & 78 & 472 & 0.8 & 0.6 & 0.66 & \textbf{0.1} & 0.49 & 0.75\\
{\color[HTML]{000000} KGS-d} & {\color[HTML]{000000} \textbf{2}} & {\color[HTML]{000000} \textbf{30}} & {\color[HTML]{000000} \textbf{43}} & {\color[HTML]{000000} \textbf{1}} & {\color[HTML]{000000} \textbf{0.89}} & {\color[HTML]{000000} \textbf{0.95}} & {\color[HTML]{000000} 0.52} & {\color[HTML]{000000} \textbf{0.3}} & {\color[HTML]{000000} \textbf{0.2}} \\
{\color[HTML]{000000} KGS-f} & {\color[HTML]{000000} 4} & {\color[HTML]{000000} 52} & {\color[HTML]{000000} 95} & {\color[HTML]{000000} 0.8} & {\color[HTML]{000000} 0.82} & {\color[HTML]{000000} 0.91} & {\color[HTML]{000000} 0.6} & {\color[HTML]{000000} 0.44} & {\color[HTML]{000000} 0.36}  \\
{\color[HTML]{000000} KGS-u} & {\color[HTML]{000000} \textbf{2}} & {\color[HTML]{000000} 36} & {\color[HTML]{000000} 105} & {\color[HTML]{000000} \textbf{1}} & {\color[HTML]{000000} 0.81} & {\color[HTML]{000000} 0.86} & {\color[HTML]{000000} 0.5} & {\color[HTML]{000000} 0.33} & {\color[HTML]{000000} 0.39} \\
{\color[HTML]{000000} KGS-c} & {\color[HTML]{000000} \textbf{2}} & {\color[HTML]{000000} 39} & {\color[HTML]{000000} 70} & {\color[HTML]{000000} \textbf{1}} & {\color[HTML]{000000} 0.87} & {\color[HTML]{000000} 0.93} & {\color[HTML]{000000} 0.51} & {\color[HTML]{000000} 0.36} & {\color[HTML]{000000} 0.3}  \\ 
 \bottomrule
\end{tabular}
\end{table*}

% \begin{figure*}[!htb]
%   \centering
%   \includegraphics[width=0.57\linewidth]{Legends.png} \\
%   \includegraphics[width=0.32\linewidth]{SHD-Syn} 
%   \includegraphics[width=0.32\linewidth]{TPR-SYN} 
%   \includegraphics[width=0.32\linewidth]{FDR-Syn} 
%   \caption{SHD (\textit{lower better}), TPR (\textit{higher better}), FDR (\textit{lower better}) plots of all the approaches on the synthetic datasets with $d$ = 10, 40, and 100.}\label{fig2}
% \end{figure*}

Table~\ref{table1} presents the performance on the synthetic datasets d-10, d-40 and d-100. From Table~\ref{table1}, it is evident that knowledge constraints have a positive impact on causal discovery due to the promising metric values. Although the extent of the impact varies for each type of knowledge, it is quite evident from the empirical results that any type of knowledge is beneficial in improving the algorithmic search quality. In terms of SHD and TPR, KGS-d does better than all other approaches for networks of all sizes. This implies that directed edges improve accuracy the most as they provide a more concise or restrained information about the causal relationship than the other constraints. Also in terms of FDR, KGS-d has the best (lowest) value for two out of three datasets. Only in case of one dataset, PC outperforms others with a lowest FDR. Overall, it seems that directed edges are more efficient at producing better graphs with less false positives and more true edges. KGS-u performs well in terms of all the metrics for the small network only. For networks of larger sizes, it does not perform significantly well. This can be due to the fact that the undecided edges provide partial information as it allows for any one of the two edge possibilities between two random variables $X_{i}$ and $X_{j}$. Either the edge direction can be from $X$ to $Y$ or $Y$ to $X$. The performance of KGS-f suggests that forbidden edges are less effective compared to constraints that provide direct information about the causal edges. This is reasonable as there are relatively %large numbers of edges absent in the ground truth graph. So, there are relatively 
more forbidden edges than the directed or undirected edges in a causal graph. The performance of KGS-c is moderately good. In terms of all three metrics, KGS-c mostly seems to have the second best results which is quite understandable as it has the combined effects of all types of edges. Mostly all the versions of KGS have better performance in case of all the datasets compared to the baseline algorithms. It indicates that any type of constraint or their combination is useful to leverage during causal graph learning.
% In terms of run time, KGS-d have the fastest speed for the datasets d-10 and d-100. For d-40, KGS-c have the best run time. It seems that as the network density increases, there is a significant difference in the run time of GES and different versions of KGS. For denser graphs (d-40 and d-100), knowledge constraints seems to help a lot in a faster convergence with good margin. 

\subsection{Real Data}\label{Real-Data}

For experimentation with real datasets, we evaluate our approach on benchmark causal graphs from the Bayesian Network Repository (BnLearn) (\cite{bnlearn}). It includes causal graphs inspired by real-world applications that are used as standards in literature. We evaluate all the versions of KGS and the baselines on three different datasets namely \textit{Child}, \textit{Alarm} and \textit{Hepar2}. Their respective ground-truth networks are available in the BnLearn repository and they vary in node and edge densities (small, medium \& large networks). The corresponding datasets are available in the \href{https://github.com/py-why/causal-learn}{causal-learn package}. We briefly introduce them below: 

\textit{(i) CHILD} (\cite{Child}) is a medical Bayesian network for diagnosing congenital heart disease in a new born "blue baby". It is a small network with $d=20$ nodes and $e=25$ edges. The dataset includes patient demographics, physiological features and lab test reports.

\textit{(ii) ALARM} (\cite{alarm}) is a healthcare application that sends cautionary alarm messages for patient monitoring. It is used to study probabilistic reasoning techniques in belief networks. The ground-truth graph is a medium sized network with $d=37$ nodes and $e=46$ edges.

\textit{(iii) HEPAR2 }(\cite{hepar2}) is a probabilistic causal model for liver disorder diagnosis. It is a Bayesian network that tries to capture the causal relationships among different risk factors, diseases, symptoms, and test results. It is a large network with $d=70$ nodes and $e=123$ edges.

\begin{table*}[!h]
\centering
\small
\setlength\tabcolsep{3.5pt} %line added to squeez space in cells
\caption{Real-world datasets results (SHD, TPR and FDR) with the best performances for each dataset boldfaced.}
\label{tab-real}
\begin{tabular}{ccccccccccccc} 
 \toprule % from booktabs package
\textbf{} & \multicolumn{3}{c}{SHD $\downarrow$} & \multicolumn{3}{c}{TPR $\uparrow$} & \multicolumn{3}{c}{FDR $\downarrow$} \\ \cmidrule(lr){2-4}\cmidrule(lr){5-7}\cmidrule(lr){8-10}
 & \textit{Child} & \textit{Alarm} &\textit{ Hepar2} & \textit{Child} & \textit{Alarm} & \textit{Hepar2} & \textit{Child }& \textit{Alarm} & \textit{Hepar2} \\  \midrule
PC & 43 & 55 & 172 & 0.24 & 0.67 & 0.35 & 0.86 & 0.6 & 0.75  \\
GES & 34 & 56 & 70 & 0.38 & 0.74 & 0.5 & 0.89 & 0.61 & 0.23  \\
dLiNGAM & 28 & \textbf{40} & 110 & 0.12 & 0.39 & 0.1 & 0.82 & \textbf{0.5} & \textbf{0.07}\\
%NOTEARS & \textbf{22} & 41 & 123 & 0.16 & 0.17 & - & \textbf{0.63} & \textbf{0.38} & -\\
KGS-d & 25 & 51 & \textbf{58} & 0.6 & \textbf{0.84} & \textbf{0.6} & 0.79 & 0.57 & 0.21 \\
KGS-f & 26 & 48 & 67 & \textbf{0.62} & 0.81 & 0.51 & 0.79 & 0.54 & 0.19  \\
KGS-u & \textbf{22 }& 52 & 59 & 0.6 & 0.82 & 0.57 & 0.78 & 0.58 & 0.2 \\
KGS-c & 23 & 47 & 61 &\textbf{ 0.62} & 0.79 & 0.55 & \textbf{0.74} & 0.55 & 0.19\\
 \bottomrule % from booktabs package
\end{tabular}
\end{table*}

% \begin{figure*}[!htb]
%   \centering
%   \includegraphics[width=0.57\linewidth]{Legends} \\
%   \includegraphics[width=0.32\linewidth]{Real_SHD.PNG} 
%   \includegraphics[width=0.32\linewidth]{Real_TPR.PNG} 
%   \includegraphics[width=0.32\linewidth]{Real_FDR.PNG} 
%   \caption{SHD (\textit{lower better}), TPR (\textit{higher better}), FDR (\textit{lower better}) plots of all the approaches on the real datasets : Child ($d = 20$), Alarm ($d = 37$), and Hepar2 ($d = 70$).}\label{fig:pitt}
% \end{figure*}

%We report the performance metrics of different versions of KGS and GES on these datasets 
Table~\ref{tab-real} present the results of real datasets. All versions of KGS performs considerably better than PC and GES in case of SHD metic. KGS-u and KGS-d have the best values of SHD for the Child and Hepar2 datasets respectively. For Alarm dataset, dLiNGAM has the best SHD.
%Compared to GES, the improvement margin in SHD for all of them is atleast by 10 which is quite impresssive. Overall, none of the SHDs of the estimated graphs by GES is better than KGS. %It seems from the SHD results that the directed ($\rightarrow$) and undirected (--) edges have a better influence in graph discovery.
In case of true positives, KGS-d seems to perform really well as it has the best TPR for two out of the three datasets (Alarm and Hepar2). For Child dataset, KGS-c and KGS-f both have the highest TPR 0.62 and the TPRs of all the versions of KGS are larger than all the baselines by a good margin. Overall, all versions of KGS have better TPRs compared to the baselines for all datasets. KGS-c outperforms others with the best FDR for the Child dataset. For rest of the datasets, dLiNGAM has the best FDRs. KGS-f and KGS-c have the second best FDR in case of Hepar2 which is still significant compared to the FDR of dLiNGAM. Because, the dLiNGAM approach performs better than KGS only twice, once w.r.t. SHD and once w.r.t FDR. Otherwise, its overall performance is mostly moderate. Especially, its TPRs are quite low. PC and GES does not achieve any best result in case of any of the real datasets. 

% The boldfaced results in the table are the ones that are better than that of KGS for the corresponding datasets. 

% In terms of run time, KGS-c is mostly faster than others with the lowest run time for two datasets (Child and Hepar2). For the other dataset Alarm, KGS-f is the fastest. One thing is common w.r.t. run time for all the 3 real datasets. The run time for KGS-f and KGS-c is quite close in all of these datasets. It signifies that forbidden edges may allow the search process to converge faster. Although forbidden edge constraints are not the best among all the edge constraints in terms of the accuracy metrics, they seem to perform well w.r.t faster convergence.

% \textit{REWRITE---
% For experimental purpose, the implementations of the algorithms have been adopted from the gCastle (\cite{gcastle} repository. From the results reported in Table~\ref{tab3}, we see that in few cases NOTEARS performs slightly better than KGS particularly for SHD and FDR metrics. However, it performs poorly in terms of discovering the true edges (TPR). Also, it could not even estimate a single graph for the case of Hepar2 dataset. 

\subsubsection{Variation of Knowledge Proportion}
\begin{figure*}[!h]
  \centering
  \includegraphics[width=0.32\linewidth]{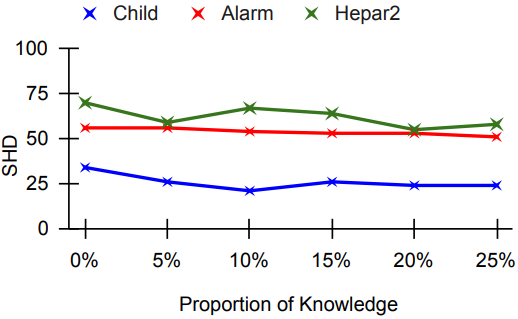} 
  \includegraphics[width=0.32\linewidth]{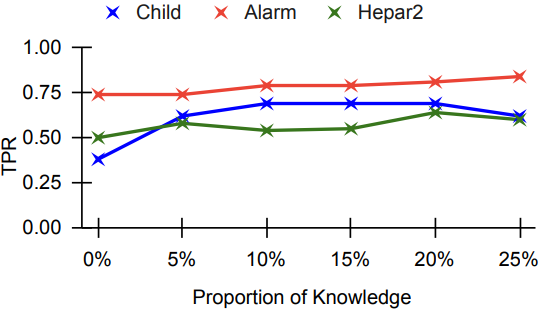} 
  \includegraphics[width=0.32\linewidth]{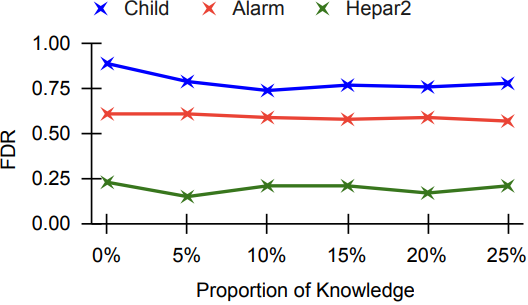} 
  \caption{Variation in SHD (\textit{lower better}), TPR (\textit{higher better}), and FDR (\textit{lower better}) due to changing the proportion of knowledge constraints in case of Child, Alarm and Hepar2 datasets.}  \label{fig:figure4}
\end{figure*}

We perform an experiment by leveraging different amount of prior edges to investigate two aspects: (i) Do raising the amount of knowledge improves the discovery? and (ii) How varying the amount of knowledge affects the structure search. This experiment is done on the real datasets by varying the amount of \textit{directed} edges from 0 to 25 percent. Each time the amount of prior knowledge is raised by 5\%. From the plots in Figure~\ref{fig:figure4}, we see that using any amount of prior knowledge is always better than using no knowledge at all (0 percent). This is because all of the metrics SHD, TPR and FDR have better values for the experiments where at least some knowledge is used. Although in some cases the improvement can be very less or same (TPR of the experiments 0 \% and 5 \% for Alarm), in no cases there is a negative impact of using knowledge. Another finding from the plots is that increasing the amount of knowledge is not always proportional to better metric values. The drop in SHD and FDR, and also, the rise in TPR due to the changing \% of knowledge is quite unsteady. However, it seems like leveraging any amount of causal knowledge is always better than leveraging no knowledge at all. Details experimental findings are listed in Appendix~\ref{appen-B}.

\subsection{Real-world Clinical Data}\label{Real-Data}

We validate our approach on a practical healthcare dataset about patients who received oxygen therapy treatment. The details of the dataset and experimental results are discussed in the following subsections.

\subsubsection{Cohort}
We examined our approach on a practical clinical application described in \cite{gani2023structural} that deals with estimating the effect of oxygen therapy (OT) on patients in intensive care units (ICU). It involves an observational dataset of about 4062 patients who received invasive mechanical ventilation (IMV) for a minimum of 168 hours or above. It followed the selection criteria and study protocols utilized during a pilot randomized control trail (RCT) as mentioned in \cite{panwar2016conservative}. The cohort selection is based on the massive MIMIC-III (\cite{johnson2016data}) critical care database which includes patients who received either liberal or conservative oxygenation treatments. The observational dataset includes a total of 26 variables, describing the ventilator settings, oxygenation measurements, and demographic information about patients. As per \cite{gani2023structural}, the ground truth graph of the dataset (see Figure~\ref{fig:OT-GT}) has 66 causal connections which they obtained relying on the majority voting of some structure learning algorithms and expert opinions. For our experiments, we adopted most of the causal connections from that study as the ground truth as well as we extracted a few more causal relations using an LLM-based information retrieval technique. The details of our analysis are mentioned in the next subsection.

\subsubsection{LLM-based Prior Knowledge Extraction}
\label{LLM-based-ext}
\begin{figure*}[!h]
  \centering
  \includegraphics[width=0.75\linewidth]{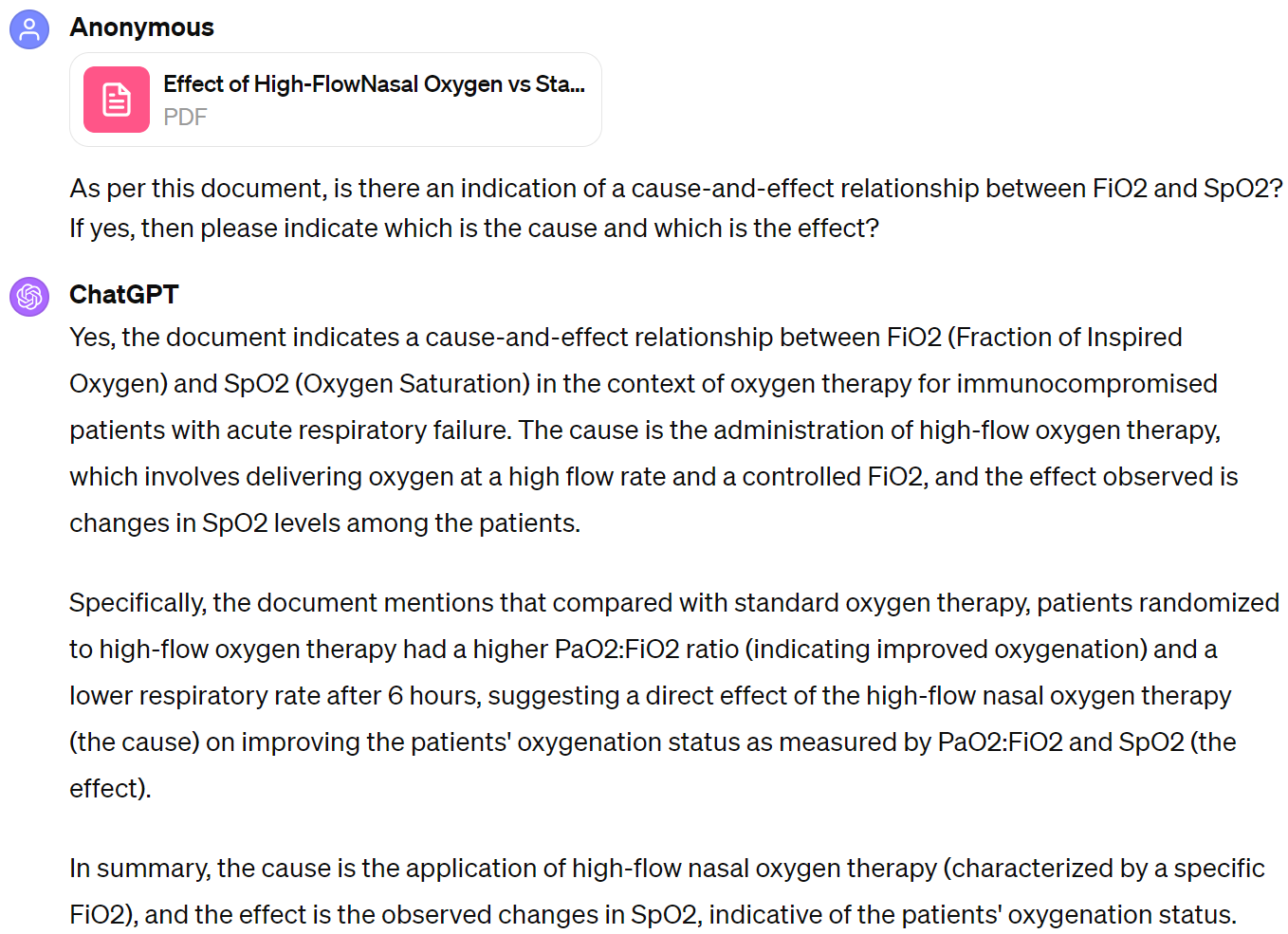} 
  \caption{Example prompt asking ChatGPT about the causal relation between two variables (FiO2 and SpO2) based on a relevant literature document. Rest of the prompts are available in Appendix~\ref{APPEN: D}.}  \label{fig:GPT promt}
\end{figure*}

Large language models (LLMs) such as GPT, BERT, etc. can be leveraged to retrieve, understand, and synthesize information from multiple literature sources. In recent years, different studies (\cite{willig2022can}, \cite{kiciman2023causal}) have explored the causal capabilities of LLMs. Prior causal knowledge can be obtained by utlizing LLMs to discover causal linkages as they are capable of analyzing large amounts of literature. In fields where experimental data or expert opinions may be scarce or difficult to obtain, such information retrieval is valuable to obtain some causal priors. In this study, we prompt the LLM GPT-4 to find the causal relationship between a given pair of variables based on a relevant literature. We collected academic papers closely related to the oxygen therapy domain and based on those, asked GPT-4 to extract causal priors if found any. Figure~\ref{fig:GPT promt} presents such an example prompt. Each of the prompt generated by the GPT-4 model \textit{has been verified }by a human expert before taking it into consideration. Based on the verfied prompts, we obtained total 17 causal edges which are listed in Table~\ref{tab:GPT-Edges}. Then we modified the OT ground truth graph based on the causal priors obtained via the LLM-based retrieval. From the 17 retrieved edges (see Table~\ref{tab:GPT-Edges}), total 3 edges were obtained newly by the LLM which we augmented with the ground truth graph. And we reversed 2 edges in the ground truth graph based on the information found by the LLM from relevant papers. To avoid errors \textit{we did not blindly rely on the LLM's answer}, rather those were verfied by an expert before incorporating. Hence, the \textit{final ground truth graph} that we used for experimentation involved total 69 edges including the newly added and reversed edges. The edges for which both GPT-4 and \cite{gani2023structural}'s ground truth graph indicate the same direction have been kept unaltered. To summarize, out of the 66 causal edges in the graph of Figure~\ref{fig:OT-GT}, we used 64 of them as it is and reversed 2 edges. Then we augmented 3 newly obtained edges which led to the total of 69 edges in our ground truth graph (see Figure~\ref{fig:New-OT-GT}). The purpose of this analysis is to study if LLMs are capable of effectively retrieving causal priors from relevant literature evidence. The answers obtained via GPT-4 supports that LLMs can be leveraged to extract causal information stated in large body of literature which can be a good source of prior knowledge for causal structure learning. All the generated prompts and papers used are available in Appendix~\ref{APPEN: D}.

\begin{table*}[!h]
\centering
\scriptsize
\caption{Extracted causal relations using GPT-4.}
\label{tab:GPT-Edges}
\begin{tabular}{cccc}
\toprule
No. & \textbf{Edge found by GPT-4} & Edge in \cite{gani2023structural} & Decision \\ \midrule
1 & PaO2 → SpO2/SpO2→PaO2 & SpO2-\textgreater{}PaO2 & Unchanged \\
2 & FiO2→SpO2 & SpO2-\textgreater{}FiO2 & Reversed \\
3 & PaO2 → SOFA & SOFA-\textgreater{}PaO2 & Unchanged \\
4 & ARDS→SpO2 & No edge in paper & Added \\
5 & PEEP→PaO2 & PEEP-\textgreater{}PaO2 & Unchanged \\
6 & COPD→SpO2 & SpO2-\textgreater{}COPD & Reversed \\
7 & COPD→FiO2 & COPD-\textgreater{}FiO2 & Unchanged \\
8 & COPD→PaCO2 & COPD-\textgreater{}PaCO2 & Unchanged \\
9 & FiO2→PaO2 & No edge in paper & Added \\
10 & Age→Trauma & Age-\textgreater{}Trauma & Unchanged \\
11 & Age→Death & Age-\textgreater{}Death & Unchanged \\
12 & SOFA→SaO2 & SOFA-\textgreater{}SaO2 & Unchanged \\
13 & SpO2→Death & SpO2-\textgreater{}Death & Unchanged \\
14 & VT→Oxygenation & VT-\textgreater{}Oxygenation & Unchanged \\
15 & Oxygenation→Death & Oxygenation-\textgreater{}Death & Unchanged \\
16 & FiO2→SaO2 & No edge in paper & Added \\
17 & COPD→PaCO2 & COPD-\textgreater{}PaCO2 & Unchanged \\ \bottomrule
\end{tabular}
\end{table*}

% \textbf{1 Para on IMP of LLM-based Priors for CD----}

\subsubsection{Results on Oxygen-therapy data}
The performance of different versions of KGS and baseline algorithms on the oxygen-therapy dataset are presented in Table~\ref{tab: OT res}. In case of the SHD metric, KGS-c outperforms all the approaches with the lowest SHD 48. Compared to the baselines, all versions of KGS in fact have a lower SHD indicating a better performance. With respect to TPR, KGS-d performs the best with the highest TPR 0.55. Compared to this value, the TPRs acheived by the baseline algorithms PC, GES and dLiNGAM is quite low. The other versions of KGS also have better TPRs than each of the baselines. KGS-c also outperforms others in terms of lower false discoveries. To summarize, KGS-c and KGS-d seems to perform significantly better than all others and KGS-f/KGS-u also have better performance metrics than each of the baseline approaches.

\begin{table*}[!h]
\centering
\scriptsize
\caption{Clinical dataset results with the best performances boldfaced. KGS-c achieved the best performance in case of two metrics: SHD and FDR, and KGS-d outperformed others with the highest TPR.}
\label{tab: OT res}
\begin{tabular}{llll}
\toprule
 & SHD $\downarrow$ & TPR $\uparrow$ & FDR $\downarrow$ \\ \midrule
PC & 73  & 0.35 & 0.69 \\
GES & 55 & 0.38 & 0.6 \\
dLiNGAM & 76 & 0.23 & 0.7 \\
KGS-d & 51 & \textbf{0.55} & 0.54 \\
KGS-f & 51 & 0.47 & 0.54 \\
KGS-u & 52 & 0.43 & 0.59 \\
KGS-c & \textbf{48} & 0.52 & \textbf{0.52}\\ \bottomrule
\end{tabular}
\end{table*}

\section{Discussion} 
Comprehending causal pathways has a significant impact on clinical decision-making, resulting in enhanced prevention strategies, customized treatment regimens, and policy formulation with the objective of enhancing public health outcomes. To optimize data-driven causal structure learning using knowledge constraints, we introduce KGS, a novel application of prior edge constraints in a score-based causal search process. We evaluate KGS's performance on a variety of synthetic and real-world datasets, including networks of different sizes and edge densities as well as on a practical clinical application. The encouraging results across multiple settings of both discrete and continuous data demonstrate the robustness and flexibility of our approach. We observe that particularly leveraging directed edges causes the best performance in most cases for improving the graphical accuracy and search. This is quite understandable as directed edges provide a complete information about the causal relation between variables. Also, the combination of edges seem to perform well. Overall, any type of edge information improves the accuracy and reliability of the graph discovery. Hence, it is evident that causal priors are helpful for guiding the search to a better trajectory and there by optimize causal discovery. We also perform an analysis that demonstrates the capability of GPT-4 to retrieve causal priors from relevant literature. This can be a powerful approach to map out causal relationships inherent in scientific texts and transform the way scholars tackle the challenging task of manual literature analysis.  
%We also demonstrate the behavior of the model and effectiveness of knowledge constraints on synthetic and real-world datasets.  
%We applied our method to both discrete and continuous data.

\paragraph{Limitations:}

There are some limitations of this study. First, in this study, we consider only causal knowledge that is completely true. That is any biased or uncertain knowledge is out of the scope of this work. Second, this approach has a limited application when there is no prior knowledge available. Although it is often the case that at least some prior knowledge about any domain can be obtained from experts or mining the literature using LLMs. In future, we want to extend this study to address biases in prior knowledge and also, study the effects of localizing the knowledge to a sub-area of the network and see how it impacts the discovery in other areas. We also plan to test how the search process can be influenced by the constraints that provide indirect guidance (soft constraints).
%\textbf{We did hard constraints only, soft constraints we can do.}

% ACKNOWLEDGEMENTS ONLY GO IN THE CAMERA-READY, NOT THE SUBMISSION
% \acks{Many thanks to all collaborators and funders!}

%Do NOT change font size of references or modify the bibliography style
\bibliography{sample}

\newpage
\appendix
\section{Background}\label{background}

\paragraph{Causal Graphical Model (CGM)}

A directed acyclic graph (DAG) is a type of graph $G$ in which the edges $e$ are directed (→) and there are no cycles. A Causal Graphical Model (CGM) consists of a DAG $G$ and a joint distribution $P$ over a set of random variables $X = (X_{1}, X_{2}, … , X_{d})$ where $P$ is Markovian with respect to $G$ (\cite{ida}). In a CGM, the nodes represent variables X, and the arrows represent causal relationships between them. The joint distribution $P$ can be factorized as follows where $pa(x_{i},G)$ denotes the parents of $x_{i}$ in $G$.\\
\begin{equation}
    P(x_{1}, …, x_{d}) =  \prod_{i=1}^n P(x_{i}|pa(x_{i},G)
    \label{eq1}
\end{equation}
A set of DAGs having the same conditional independencies belong to the same equivalence class. 
DAGs can come in a variety of forms based on the kinds of edges they contain. A Partially Directed Graph (PDAG) contains both directed and undirected edges. A Completed PDAG (CPDAG) consists of directed edges that exist in every DAG $G$ belonging to the same equivalence class and undirected edges that are reversible in $G$. %DAGs that belong to the same equivalence class have identical set of independencies. 

\paragraph{Score-based Causal Discovery}

A score-based causal discovery approach typically searches over the equivalence classes of DAGs to learn the causal graph $G$ that best fits the observed data $D$ as per a score function $S(G,D)$ which returns the score $S$ of $G$ given data $D$ (\cite{chickering2002optimal}, \cite{k-notears}). Here, the optimization problem for structure learning is as follows:
%The algorithm searches over the equivalence classes of DAGs $H$ where $G$ $\in$ $H$
\begin{equation}
    \underset{ \text{subject to } G \in D}{ \mathop{\min}_{G}} S(G,X) 
    \label{optimization}
\end{equation}

Typically, any score-based approach has two main components:\textit{ (i) a search strategy} - to traverse the search space of candidate graphs\textit{ $G$, and (ii) a score function - }to evaluate the candidate causal graphs.

\paragraph{Score Function} A scoring function  $S(G,D)$  maps causal DAGs $G$ to a numerical score, based on how well $G$ fits to a given dataset $D$. A commonly used scoring function to select causal models is the Bayesian Information Criterion (BIC) (\cite{BIC}) which is defined below: 

\begin{equation}
    S_{BIC} = -2 * loglikelihood + k * log(n),
\end{equation}

where $n$ is the sample size used for training and $k$ is the total number of parameters. %The lower BIC score signals a better model.

\section{Additional Simulation Results}\label{appen-B}

\paragraph{Experimental results of varying the knowledge proportion}

We present the details of all the metric values for the experiment done with varying the amount of prior knowledge in Table~\ref{tab2}. This experiment is done by varying the amount of constraints (directed edges) from 0 to 25 percent each time by raising the amount of knowledge by 5\%. The results show that any amount of knowledge is good for improving search accuracy and hence should be leveraged during the search process. Although it is surprising that the increment in knowledge is not directly proportional to the increment in discovery accuracy. Still leveraging any percentage of knowledge is better than using no knowledge at all.
%Proportion of knowledge experiment

% \usepackage{multirow}
\begin{table*}[!h]
\small
\centering
\caption{Results (metric values) of the experiment done by varying knowledge proportion.}
\label{tab2}
\begin{tabular}{|c|c|c|c|c|c|}
 \toprule
Datasets & \begin{tabular}[c]{@{}c@{}}Proportion of \\ Knowledge\end{tabular} & \begin{tabular}[c]{@{}c@{}}SHD\\ (lower better)\end{tabular} & \begin{tabular}[c]{@{}c@{}}TPR \\ (higher better)\end{tabular} & \begin{tabular}[c]{@{}c@{}}FDR \\ (lower better)\end{tabular} & \begin{tabular}[c]{@{}c@{}}Estimated models \\ (lower better)\end{tabular} \\ \hline
Child & 0\% & 34 & 0.38 & 0.89 & 49 \\
 & 5\% & 26 & 0.62 & 0.79 & 39 \\
 & 10\% & \textbf{21} & \textbf{0.69} & \textbf{0.74} & 36 \\
 & 15\% & 26 & \textbf{0.69} & 0.77 & 37 \\
 & 20\% & 24 & \textbf{0.69} & 0.76 & 35 \\
 & 25\% & 24 & 0.62 & 0.78 & \textbf{34} \\ \hline
Alarm & 0\% & 56 & 0.74 & 0.61 & 84 \\
 & 5\% & 56 & 0.74 & 0.61 & 82 \\
 & 10\% & 54 & 0.79 & 0.59 & 83 \\
 & 15\% & 53 & 0.79 & 0.58 & 76 \\
 & 20\% & 53 & 0.81 & 0.59 & 77 \\
 & 25\% & \textbf{51} & \textbf{0.84} & \textbf{0.57} & \textbf{74} \\ \hline
Hepar2 & 0\% & 70 & 0.5 & 0.23 & 83 \\
 & 5\% & 59 & 0.58 & \textbf{0.15} & 81 \\
 & 10\% & 67 & 0.54 & 0.21 & 76 \\
 & 15\% & 64 & 0.55 & 0.21 & 73 \\
 & 20\% & \textbf{55} & \textbf{0.64} & 0.17 & 71 \\
 & 25\% & 58 & 0.6 & 0.21 & \textbf{66} \\
 \bottomrule
\end{tabular}
\end{table*}

\section{Baseline Causal Discovery Approaches}

We report the performance of different baseline causal discovery approaches such as PC (constraint-based), GES (score-based) and LiNGAM (FCM-based) on the experimental datasets to see their comparative performance with respect to KGS. We briefly discuss the methods below:

\textbf{(i) PC:}
The Peter-Clark (PC) algorithm (\cite{PC}) is a very common constraint-based causal discovery approach that largely depends on conditional independence (CI) tests to find the underlying causal graph. Primarily, it works in three steps: (i) Skeleton construction, (ii) V-structures determination, and (iii) Edge orientations.

\textbf{(ii) GES: } Greedy Equivalence Search, GES (\cite{chickering2002optimal}) is one of the oldest score-based causal discovery methods that employ a greedy search over the space of equivalence classes of DAGs. 
% Each search state is represented by a CPDAG where edge modification operators such as insert and delete operators allow for single-edge additions and deletions respectively. 
%After the an edge modification to the current CPDAG, a score function is used to score the model. If the new score is better than the current score, only then the modification is allowed. 
Primarily GES operates in two phases: (i) Forward Equivalence Search (FES) and (ii) Backward Equivalence Search (BES). 
% The first phase FES starts with an empty (i.e., no-edge) CPDAG, and greedily adds edges by considering every single-edge addition that could be performed to every DAG $G$ in the current equivalence class. This phase continues until it reaches a local maximum. After that the second phase BES starts where at each step, it considers all possible single-edge deletions. This continues until there is an improvement of the score. Finally, GES terminates once the second phase reaches a local maximum. 
GES assumes a decomposable score function $S(G,D)$ which is expressed as a sum of the scores of individual nodes and their parents. A problem with GES is that the number of search states that it needs to evaluate scales exponentially with the number of nodes $d$ in the graph (\cite{SGES}). This results in a vast search space, and also scoring a large number of graphs which adds to the overall cost as score computation is an expensive step.
% The branching factor of the search space can grow exponentially w.r.t the number of nodes if the models reached by FES are complex (\cite{SE-GES}). This makes GES almost impractical to use on large, complex problems.

\begin{equation}
    S(G,D) = \sum_{i=1}^{d} s(x_{i}, pa(x_{i},G))
\end{equation}

\textbf{(iii) LiNGAM: }Linear Non-Gaussian Acyclic Model (LiNGAM) uses a statistical method known as independent component analysis (ICA) to discover the causal structure from observational data. It makes some strong assumptions such as the data generating process is linear, there are no unobserved confounders, and noises have non-Gaussian distributions with non-zero variances (\cite{LiNGAM}). 

\textbf{DirectLiNGAM} (dLiNGAM) is an efficient variant of the LiNGAM approach that uses a direct method for learning a linear non-Gaussian structural equation model (\cite{directlingam}). The direct method estimates causal ordering and connection strengths based on non-Gaussianity.

\section{Performance Metrics}

• \textbf{Structural Hamming Distance (SHD):} SHD is the sum of the edge additions (A), deletions (D) or reversals (R) that are required to convert the estimated graph into the true causal graph (\cite{notears, cheng2022evaluation}). To estimate SHD it is required to determine the missing edges, extra edges and edges with wrong direction in the estimated graph compared to the true graph. Lower the SHD closer is the graph to the true graph and vice versa. The formula to calculate SHD is given below:

\begin{equation}
    SHD = A + D + R
\label{eq1}
\end{equation}

• \textbf{True Positive Rate (TPR):} TPR denotes the proportion of the true edges in the actual graph that are correctly identified as true in the estimated graph. A higher value of the TPR metric means a better causal discovery.

\begin{equation}
    TPR = \frac{TP}{TP+FN}
\end{equation}

Here, TP means the true positives or the number of correctly identified edges and FN or false negatives denote the number of unidentified causal edges.

• \textbf{False Discovery Rate (FDR):} FDR is the ratio of false discoveries among all discoveries (\cite{notears}). FDR represents the fraction of the false edges over the sum of the true and false edges. Lower the FDR, better is the outcome of causal discovery. 

%\autoref{fdr} is used to compute FDR where $FP$ = total number of false positives and $TP$ = total number of true positives.

\begin{equation}
     FDR = \frac{FP}{TP+FP}
\label{fdr}
\end{equation}

Here, FP or false positives represent the number of wrongly identified directed edges.

\section{Extraction of Causal Priors using LLMs}
\label{APPEN: D}

% List of the extracted causal edges along with their corresponding prompts and answers:\\

\begin{table}[!h]
\centering
\small
\caption{The causal relationships retrieved by GPT-4 from relevant literature papers and their corresponding prompt-answer link.}
\label{tab:my-table}
\begin{tabular}{|c|c|}
\hline
\textbf{Edge by GPT-4} & \textbf{Chat Link of GPT-4 Prompts and Answers} \\ \hline
PaO2 → SpO2 &  \url{https://chat.openai.com/share/5dc1db93-b394-4bcd-a5de-837d0863e31f} \\ \hline
SpO2→PaO2 & \url{https://chat.openai.com/share/cff9f68c-4bb7-4f23-86a8-056aa7fd1955} \\ \hline
FiO2→SpO2 & \url{https://chat.openai.com/share/1b3a5e3c-d84d-41d7-a2c4-263ebb6766f6} \\ \hline
PaO2 → SOFA & \url{https://chat.openai.com/share/07fa3389-e8ec-41f2-b45e-71c0496c7852} \\ \hline
ARDS→SpO2 & \url{https://chat.openai.com/share/31644194-7aed-4ad5-95ae-bcfc4c8ec6d1} \\ \hline
PEEP→PaO2 & \url{https://chat.openai.com/share/7dccd77e-3e0d-4d05-9624-2957148c4490} \\ \hline
COPD→SpO2 & \url{https://chat.openai.com/share/6fb4857c-1656-402e-8eae-d56f6c2f8c20} \\ \hline
COPD→FiO2 & \url{https://chat.openai.com/share/b576b2eb-4d73-484e-9cac-99fc02bcbe92} \\ \hline
COPD→PaCO2 & \url{https://chat.openai.com/share/32c83464-9498-42b6-bd15-2f94fb2d0d9c} \\ \hline
FiO2→PaO2 & \url{https://chat.openai.com/share/66c04507-a981-42a8-8d89-f56eff9521c4} \\ \hline
Age→Trauma & \url{https://chat.openai.com/share/a49042e8-ea48-42a6-a0c2-2ae1b4b9492e} \\ \hline
Age→Death & \url{https://chat.openai.com/share/daca5caf-48b7-4168-988a-7ae3e44ccaf8} \\ \hline
SOFA→SaO2 & \url{https://chat.openai.com/share/43659587-c1da-4714-a262-1194671cc9c5} \\ \hline
SpO2→Death & \url{https://chat.openai.com/share/ee3ce03a-7951-4999-9dd2-395665871838} \\ \hline
VT→Oxygenation & \url{https://chat.openai.com/share/57905bad-9972-4e37-8f1b-912127ebd226} \\ \hline
Oxygenation→Death & \url{https://chat.openai.com/share/79134416-c2fa-476b-b790-bf9e8bab7d89} \\ \hline
FiO2→SaO2 & \url{https://chat.openai.com/share/cc09e77b-589e-4ee9-a4db-5add3d81b69a} \\ \hline
COPD→PaCO2 & \url{https://chat.openai.com/share/919b0fa4-ea06-4097-8578-058303852941} \\ \hline
\end{tabular}
\end{table}

%---ADD TABLE OF 10 PAPERS---

\textbf{List of the academic papers} relevant to oxygen therapy dataset used by GPT-4:

\begin{enumerate}
    \item Does age matter? The relationship between age and mortality in penetrating trauma (\cite{ottochian2009does}).
    \item Conservative versus Liberal Oxygenation Targets for Mechanically Ventilated Patients A Pilot Multicenter Randomized Controlled Trial (\cite{panwar2016conservative}).
    \item Conservative Oxygen Therapy during Mechanical Ventilation in the ICU (\cite{icu2020conservative}).
    \item Effect of Conservative vs Conventional Oxygen Therapy on Mortality Among Patients in an Intensive Care Unit The Oxygen-ICU Randomized Clinical Trial (\cite{girardis2016effect}).
    \item Liberal or Conservative Oxygen Therapy for Acute Respiratory Distress Syndrome (\cite{barrot2020liberal}).
    \item Effect of High-Flow Nasal Oxygen vs Standard Oxygen on 28-Day Mortality in Immunocompromised Patients With Acute Respiratory Failure The HIGH Randomized Clinical Trial (\cite{azoulay2018effect}).
    \item Oxygen Therapy in Chronic Obstructive Pulmonary Disease (\cite{kim2008oxygen}).
    \item Effect of Low-Normal vs High-Normal Oxygenation Targets on Organ Dysfunction in Critically Ill Patients A Randomized Clinical Trial (\cite{gelissen2021effect}).
    \item Oxygen-Saturation Targets for Critically Ill Adults Receiving Mechanical Ventilation (\cite{semler2022oxygen}).
    \item Mechanical Ventilation: State of the Art (\cite{pham2017mechanical}).
\end{enumerate}

To avoid any mistakes we did not blindly rely on the LLM’s statements, rather we verfied those by an expert before incorporating in the ground truth graph.

\section{Ground Truth Graphs}

\begin{figure*}[!h]
  \centering
  \includegraphics[width=1\linewidth]{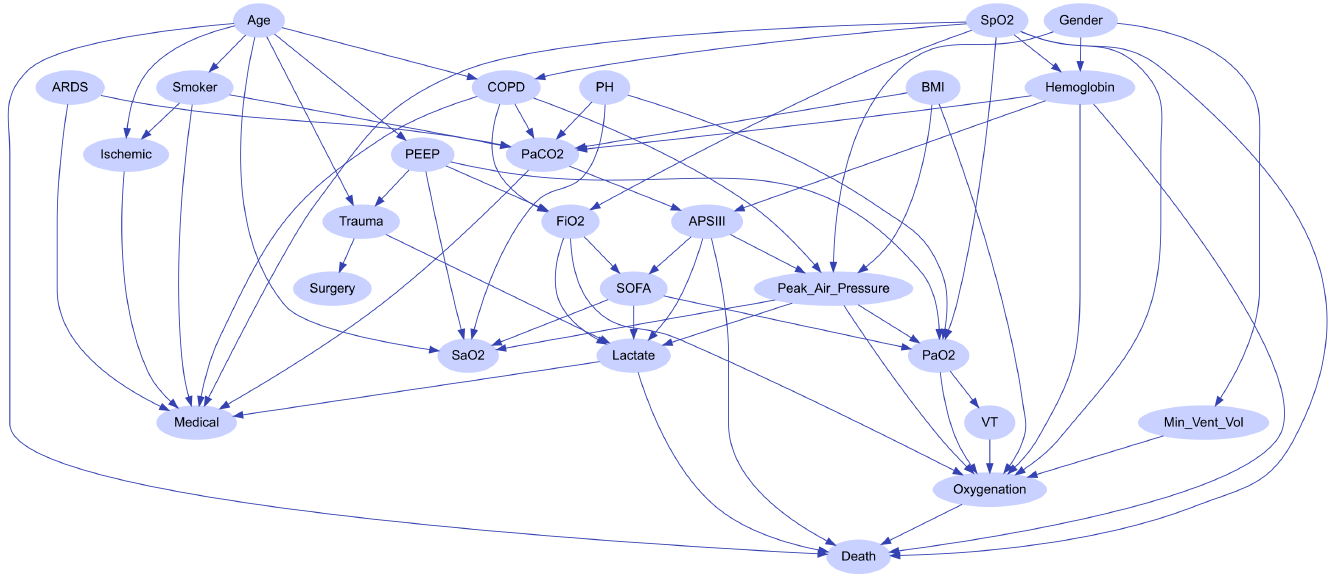} 
  \caption{Ground truth graph of Oxygen Therapy dataset from \cite{gani2023structural}.}  \label{fig:OT-GT}
\end{figure*}

\begin{figure*}[!h]
  \centering
  \includegraphics[width=1\linewidth]{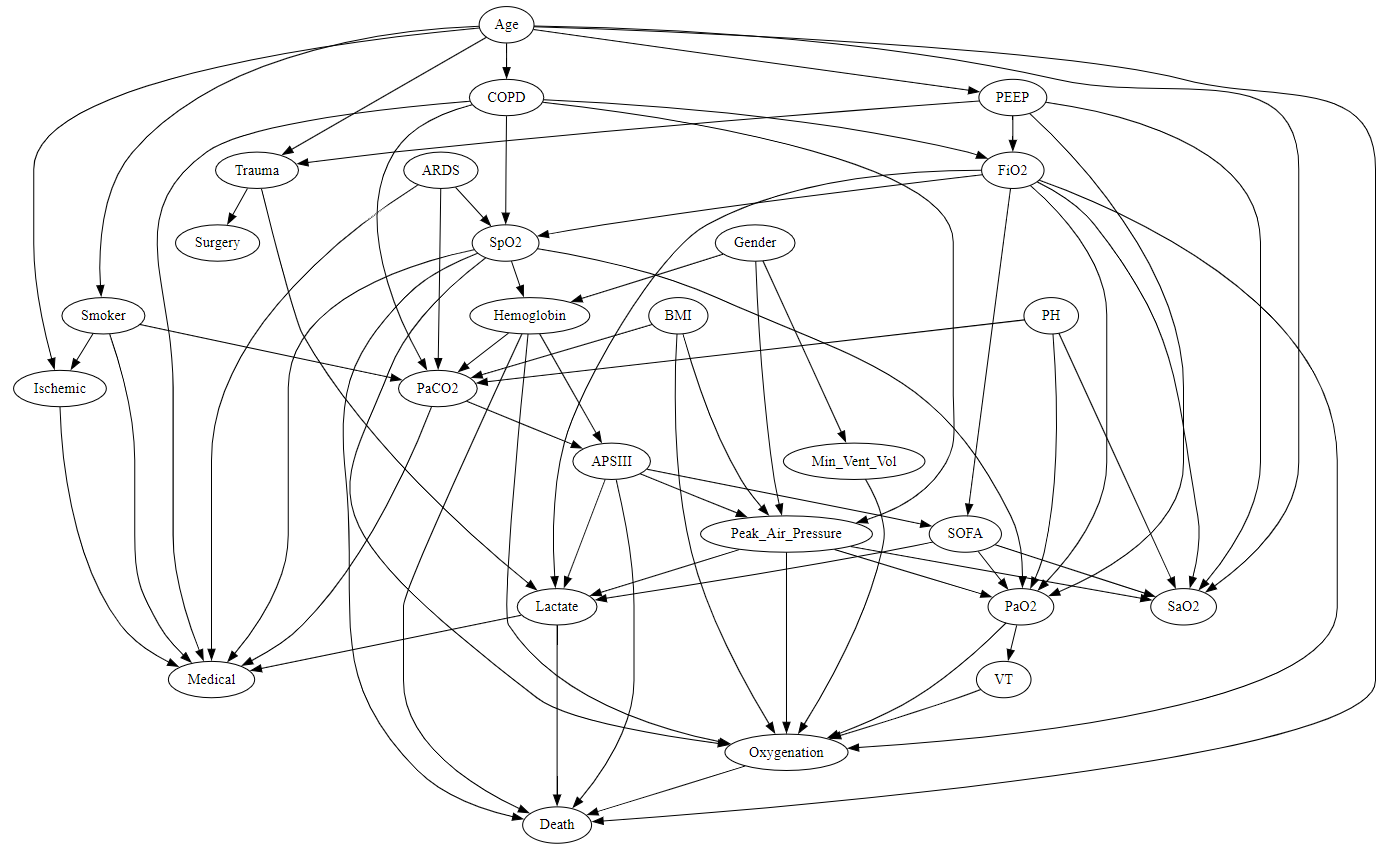} 
  \caption{Reformed ground truth graph of Oxygen Therapy dataset with 69 causal edges that we used for our experiments. Here, the reversed edges are: FiO2→SpO2 and COPD→SpO2. The added edges are ARDS→SpO2, FiO2→PaO2 and FiO2→SaO2.}  \label{fig:New-OT-GT}
\end{figure*}

\paragraph{Oxygen Therapy $G_{T}$} Figure~\ref{fig:OT-GT} represents the ground truth graph of the OT dataset available in the study \cite{gani2023structural}. This graph has total 66 causal edges. We used 64 of them exactly, reversed 2 edges and augmented 3 newly obtained edges. The 2 reversed edges are: FiO2→SpO2 and COPD→SpO2. The 3 added edges are ARDS→SpO2, FiO2→PaO2 and FiO2→SaO2. Other than these modifications, all other edges in Figure~\ref{fig:OT-GT} have been kept unaltered in the ground truth that we used (see Figure~\ref{fig:New-OT-GT}) for our experiments.

\paragraph{Child $G_{T}$} The ground truth graph of the Child network is available in \url{https://www.bnlearn.com/bnrepository/discrete-medium.html#child}.

% \begin{figure*}[!h]
%   \centering
%   \includegraphics[width=0.8\linewidth]{Child-GT.png} 
%   \caption{Ground truth graph of Child dataset from BnLearn repository (\cite{bnlearn}).}  \label{fig:Ch-GT}
% \end{figure*}

\paragraph{Alarm $G_{T}$} The ground truth graph of the Alarm network is available in \url{https://www.bnlearn.com/bnrepository/discrete-medium.html#alarm}.

% \begin{figure*}[!h]
%   \centering
%   \includegraphics[width=0.8\linewidth]{Alarm-GT.png} 
%   \caption{Ground truth graph of Alarm dataset from BnLearn repository (\cite{bnlearn}).}  \label{fig:OT-GT}
% \end{figure*}

\paragraph{Hepar2 $G_{T}$} The ground truth graph of the Hepar2 network is available in \url{https://www.bnlearn.com/bnrepository/discrete-large.html#hepar2}.

\section{Code Availability}

After the blind review period is over, we will add a link to a public repository for the code and datasets. For now, we have uploaded the code and datasets of KGS for review as supplementary material.

\end{document}